\documentclass[a4paper]{article}
\pdfoutput=1

\usepackage{INTERSPEECH2022}

\usepackage{amsmath,amsfonts,amssymb}
\usepackage{mathtools}
\usepackage{microtype}
\usepackage{booktabs}
\usepackage{graphicx}
\usepackage{hyperref}
\usepackage{url}
\usepackage{xcolor}

\usepackage{pdflscape}
\usepackage{cite}

\usepackage[font=small,skip=2ex]{caption}

\usepackage{multirow}
\usepackage{etoolbox}

\let\citet\cite
\let\citep\cite

\newrobustcmd{\B}{\bfseries}

\title{Leveraging unsupervised and weakly-supervised data to\\ improve direct speech-to-speech translation}
\name{Ye Jia,~~Yifan Ding,~~Ankur Bapna,~~Colin Cherry,~~Yu Zhang,~~Alexis Conneau,~~Nobuyuki Morioka}
\address{Google Research}
\email{\{jiaye,dyf,ankurbpn,colincherry,ngyuzh,aconneau,nmorioka\}@google.com}

\begin{document}

\maketitle

\begin{abstract}
End-to-end speech-to-speech translation (S2ST) without relying on intermediate text representations is a rapidly emerging frontier of research. Recent works have demonstrated that the performance of such direct S2ST systems is approaching that of conventional cascade S2ST when trained on comparable datasets. However, in practice, the performance of direct S2ST is bounded by the availability of paired S2ST training data. In this work, we explore multiple approaches for leveraging much more widely available unsupervised and weakly-supervised speech and text data to improve the performance of direct S2ST based on Translatotron 2. With our most effective approaches, the average translation quality of direct S2ST on 21 language pairs on the CVSS-C corpus is improved by +13.6 BLEU (or +113\% relatively), as compared to the previous state-of-the-art trained without additional data. The improvements on low-resource language are even more significant (+398\% relatively on average). Our comparative studies suggest future research directions for S2ST and speech representation learning.
\end{abstract}

\section{Introduction}

Speech-to-speech translation (S2ST) aims to break the communication barriers between people speaking different languages. Conventionally, S2ST systems are built with a cascade of automatic speech recognition (ASR), text-to-text machine translation (MT), and text-to-speech synthesis (TTS) sub-systems, all of which rely on intermediate text representations. Recently, there has been progress towards developing S2ST without relying on intermediate text representations, such as end-to-end direct S2ST \citep{jia2019direct,kano2021transformer,jia2021translatotron} and cascaded S2ST based on discrete speech representations \citep{tjandra2019speech,zhang2020uwspeech,lee2021direct,ma2021direct,lee2021textless}.
These works have demonstrated several advantages over conventional cascade S2ST systems, such as preserving paralinguistic and non-linguistic speech information during translation (e.g. speakers' voices) \citep{jia2019direct,jia2021translatotron}, supporting languages without written form \citep{tjandra2019speech,zhang2020uwspeech,lee2021direct,lee2021textless}, avoiding error compounding across sub-systems and better handling of content that does not need to be translated (e.g. names and proper nouns) \citep{jia2019direct,jia2022cvss}. %

However, training direct S2ST models without relying on intermediate text typically requires a large amount of parallel S2ST training data, which is scarce and significantly more expensive to obtain than acquiring data for training each individual component in a conventional cascade S2ST system.
While some of these challenges are shared with research on direct speech-to-text translation (ST), there are several challenges unique to S2ST. To train a S2ST model that generates translated speech with high naturalness and cleanness, it is usually desirable to collect clean target speech with high acoustic quality, which can significantly drive up the cost of data collection. 
Although it has been demonstrated in \citet{jia2022cvss} that the performance of direct S2ST is approaching cascade S2ST with a very small gap, when the two are trained on comparable datasets, the lack of large scale high-quality training data is still a primary limiting factor for direct S2ST models in practice.

In this paper, we first propose a small change to the Translatotron~2 S2ST model \citep{jia2021translatotron}, which significantly improves its translation quality on the CVSS\nobreakdash-C corpus \citep{jia2022cvss} by +1.4 BLEU (or +16\% relatively) average on 21 language pairs.
Secondly, we conduct empirical studies exploring multiple approaches to leverage more widely available unsupervised data (e.g. unlabeled speech and text) and weakly-supervised data (e.g. labeled ASR and MT data) to improve the translation quality of Translatotron~2. With the most effective approaches applied, we further improve the translation quality on the CVSS-C corpus by +15.5 BLEU (or +153\% relatively). Our best result outperforms the prior state-of-the-art, which was trained only on CVSS-C, by +13.6 BLEU (or +113\% relatively) on average on 21 language pairs, and +398\% relatively on average on 12 low-resource language pairs.
Our comparative studies suggest future research directions on S2ST and speech representation learning.

\section{Related works}

In the past few years, self-supervised speech representation learning has been shown very successful on ASR \cite{oord2018representation,baevski2020wav2vec,hsu2021hubert,chung2021w2v} and ST \cite{wu2020self,li2020multilingual,wang2021large,anastasopoulos2021findings,babu2021xlsr} tasks. Recently, it has been expanded to also learn from text, with the emergence of semi-supervised speech-text joint representation learning \citep{zheng2021fused,bapna2021slam,bapna2022mslam}. The success of these self-supervised methods has been magnified by scaling up the size of the training data and the models \citep{babu2021xlsr,bapna2022mslam}. In fact, the progress on such representation learning on massive amounts of unlabeled data has been the biggest driver for performance improvements in direct ST on a few benchmarks, e.g. CoVoST~2 \citep{wang2020covost2}.

Before the emergence of self-supervised speech representation learning, direct ST often leveraged pre-training on other supervised tasks, e.g. ASR and MT, in order to benefit from additional weakly-supervised data \citep{berard2018end,bansal2018pre}.
Other approaches for utilizing weakly-supervised data in direct ST include multi-task learning \citep{weiss2017sequence,anastasopoulos2018tied}, knowledge distillation \citep{jia2019leveraging,liu2019end}, and TTS-based data augmentation using a back-translation style approach \citep{jia2019leveraging} .

Although most of the above approaches can be expanded to apply to direct S2ST, there has been little work on examining their effectiveness. \citet{jia2022cvss} explored pre-training part of the direct S2ST model on weakly-supervised tasks, e.g. ASR and ST, which brought the performance of direct S2ST close to cascade S2ST trained on comparable datasets.
\citet{lee2021direct} built cascade S2ST based on discrete speech representations learned from a large unlabeled speech dataset, but its performance was not better than recent direct S2ST models trained from scratch, e.g. \citep{jia2021translatotron}.
\citet{duquenne2021multimodal,lee2021textless} demonstrated that it is possible to mine S2ST data from multilingual untranscribed speech corpora and improve the performance of the trained S2ST models.

\section{Methods}
\label{sec:method}

\subsection{Direct S2ST model}
\label{sec:model}

Translatotron 2 \citep{jia2021translatotron} is a direct S2ST model that can be trained end-to-end. It consists of a speech encoder, a linguistic decoder, an acoustic synthesizer, and a single attention module that connects them together. The use of a single attention module provides temporally synchronized linguistic information (in the translated language) and acoustic information (from the source speech) to the synthesizer. As a result, the model is capable of preserving fine granularity non-linguistic and paralinguistic information during translation.

The original Translatotron 2 uses a Conformer \citep{gulati2020conformer} encoder, an LSTM decoder, and a duration-based LSTM synthesizer. In this work, we modify it to use an autoregressive Transformer decoder \citep{vaswani2017attention}, which obtained better translation quality in our experiments (Sec.~\ref{exp:baselines}).

\subsection{Pre-training}
\label{sec:pre-trainng}

We explore multiple pre-training strategies in order to utilize large amounts of unsupervised or semi-supervised speech and/or text data, as well as weakly-supervised MT data for improving the performance of direct S2ST.

\subsubsection{Self-supervised speech representation learning}
\label{sec:w2v-bert}

To utilize large amounts of unsupervised speech data, we use speech representations learned from a w2v-BERT \citep{chung2021w2v} model. 

w2v-BERT is a self-supervised learning model that combines contrastive learning and masked language model (MLM). It consists of three components which are trained end-to-end: a feature encoder that encodes input speech signals into a continuous latent representation; a contrastive network that discretizes continuous latent representations into discrete speech tokens, optimized by a contrastive objective; and a context network that learns contextualized speech representations on the discretized speech tokens by optimizing an MLM objective.

We pre-train a w2v-BERT model on approximately 429k hours of unlabeled speech data in 51 languages. This dataset is originated from a combination of VoxPopuli~\citep{wang2021voxpopuli}, Common Voice~\citep{ardila2020common}, MLS~\citep{pratap2020mls} and Babel~\citep{gales2014speech}, as described in \citet{bapna2022mslam}. The pre-trained w2v-BERT model is used as the encoder of Translatotron 2 and is fine-tuned on its S2ST task. %

\subsubsection{Semi-supervised speech-text joint representation learning}
\label{sec:mslam}

In addition to unsupervised speech data, we also explore using a large amount of unsupervised text data and a small amount of paired speech-transcript data in a semi-supervised manner. To this end, we use speech-text joint representations learned from an mSLAM \citep{bapna2022mslam,bapna2021slam} model. One advantage of this speech-text joint representation is the ability to fine-tune these models in a multi-task setup (Sec.~\ref{sec:multitask}), to make use of text-based weakly-supervised data, such as MT data.

mSLAM combines w2v-BERT and SpanBERT \citep{joshi2020spanbert} into a single model to support speech-text joint representation learning. It is trained in a multi-task learning setup that can learn from unlabeled speech, unlabeled text, and paired speech-transcript data. The training on unlabeled speech is identical to w2v-BERT. For unlabeled text, it prepends a text embedding layer to w2v-BERT's context network  to form a text encoder that is trained with a SpanBERT objective. To encourage alignment between the learned speech and text representations, it adds a Connectionist Temporal Classification (CTC) objective \citep{graves2006connectionist} and a Translation Language Model (TLM) objective \citep{conneau2019cross}, which are trained on a small amount of speech-transcript paired data.

We pre-train a mSLAM model on the same unlabeled speech data as described in Sec.~\ref{sec:w2v-bert}, plus the mC4 \citep{xue2020mt5} unlabeled text corpus containing 6T tokens in 101 languages, and 2.4k hours of paired speech-transcript data spanning 32 languages, as described in \citet{bapna2022mslam}.
The pre-trained mSLAM model is used as the speech encoder in Translatotron 2 and the text encoder in MT models, and is fine-tuned on the corresponding tasks.

\subsubsection{Supervised sequence-to-sequence pre-training}

S2ST and MT share a lot of similarity in terms of translating linguistic concepts from one language to another. To utilize the more widely available MT data in the S2ST task, we explore pre-training the linguistic decoder of Translatotron 2 with MT tasks on a large MT dataset.
Because the linguistic decoder of Translatotron~2 predicts phonemes instead of text tokens, and its input is encoded speech representations instead of text, we explore multiple variants of MT tasks that predict phonemes as output and/or take phonemes as inputs.

We pre-trained Transformer-based MT models on a large MT dataset including 21 X$\to$English language pairs. This dataset is originated from WMT \citep{barrault2020findings,barrault2019findings,bojar2018findings,bojar2017findings,bojar2015findings,bojar2013findings}, TED \citep{qi2018and} and CoVoST 2 \citep{wang2020covost2}, as described in \citet{bapna2022mslam}.
The decoder of the pre-trained MT model is used for initializing the linguistic decoder of Translatotron 2.

\begin{figure}[t]
  \centering
  \includegraphics[width=0.6\linewidth]{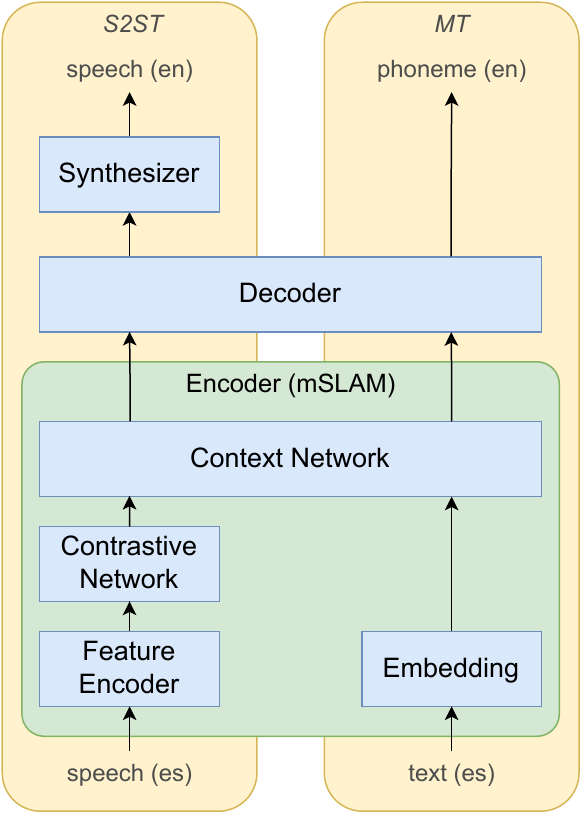}%
  \caption{S2ST and MT multimodal multi-task fine-tuning with a shared pre-trained mSLAM encoder and a shared decoder.}
  \label{fig:multitask-finetuning}
\end{figure}

\subsection{Multimodal multi-task fine-tuning}
\label{sec:multitask}

While pre-training the decoder in an MT task can potentially transfer the knowledge learned from the MT data to the S2ST model, such transferred knowledge may be susceptible to catastrophic forgetting during the course of the fine-tuning~\citep{kirkpatrick2017overcoming}. To mitigate this potential degradation, we explore fine-tuning the Translatotron 2 model in a S2ST and MT multi-modal multi-task learning setup, with the decoder shared by the two tasks.
For consistency between the two tasks, we modify the MT task to predict phonemes in the target language instead of text tokens.
When the pre-trained speech-text joint encoder (Sec.~\ref{sec:mslam}) is used, the context network in the encoder can be shared and fine-tuned on both tasks as well.
Figure~\ref{fig:multitask-finetuning} illustrates the
S2ST and MT
multimodal multi-task fine-tuning with a shared pre-trained encoder and a shared decoder, in which unsupervised speech and text data, speech-transcript pair data, and weakly-supervised MT data are all utilized.

\subsection{TTS-based data augmentation}

As an alternative to utilizing weakly-supervised MT data, we use a multilingual multi-speaker TTS model to augment the MT data into synthetic S2ST data, and train the Translatotron 2 model in a mixture of the original S2ST dataset and the augmented dataset. Compared to the multimodal multi-task fine-tuning approach, this method has the advantage of training the model entirely in the matched modality and thus potentially better utilizing the weakly-supervised data. On the other hand, one possible shortcoming is that it may bias the trained model towards synthetic speech input but not generalize well to human speech. As \citep{jia2019leveraging} suggests, such concerns can be addressed by jointly fine-tuning on a mixture of real and synthetic data.

\section{Experiments}

To compare the effectiveness of the various approaches described in Sec.~\ref{sec:method}, we conducted comparative experiments on the CVSS-C corpus~\citep{jia2022cvss}. CVSS-C is derived from the CoVoST~2 and Common Voice corpora, containing sentence-level paired S2ST data in 21 X$\to$English language pairs. The source speech in the corpus is 1,153 hours of human read speech collected via crowdsourcing; the target speech in the corpus is 719 hours of high-quality TTS synthetic speech, with naturalness on-par with human recordings. The target speech is shorter than the source speech because of better fluency in the TTS synthetic speech.

We used the modified Translatotron 2 (Sec.~\ref{sec:model}) as the direct S2ST model, implemented in the Lingvo framework \citep{shen2019lingvo}. Unless specified otherwise, all the models followed the hyper-parameters from \citet{jia2022cvss} except for the decoder, which used a 512$\times$6 Transformer following hyper-parameters from \citep{vaswani2017attention}. The decoder used a higher dropout probability 0.3 when trained or fine-tuned on CVSS-C only; otherwise the default probability 0.1 was used.

Following \citet{jia2019direct,jia2022cvss}, the translation quality of S2ST is evaluated by BLEU on ASR transcription from the translation speech (in lowercase, excluding punctuation marks). We used the same ASR model  from \citep{park2020improved} for evaluation, following \citet{jia2022cvss}, therefore the results are comparable to \citet{jia2022cvss}. The results are grouped into high/mid/low-resource language groups based on the amount of data available in the CVSS-C corpus, following \citet{babu2021xlsr,bapna2022mslam}.

\begin{table}[t]
  \small
  \caption{Performance of different approaches for utilizing unsupervised, semi-supervised, and weakly-supervised data in direct S2ST, measured by average ASR-BLEU in multilingual X$\to$En S2ST on the CVSS-C corpus, grouped into high/mid/low-resource language pairs. All models used decoders/synthesizers with 25M/27M parameters. Pre-trained/non-pre-trained encoders had 0.6B/26M parameters.
  }
  \label{tbl:comparison}
  \centering
\begin{tabular}{l@{\hskip 2.5em}rrrr}
    \toprule
    & All & High & Mid& Low \\
    \midrule
    \multicolumn{5}{l}{\kern-0.5em\emph{From scratch}} \\
    Baseline \citep{jia2022cvss} &  8.7 & 25.4 & 12.6 & 1.5 \\
    This work                    & 10.1 & 26.9 & 14.2 & 2.8 \\
    \midrule
    \multicolumn{5}{l}{\kern-0.5em\emph{Encoder pre-training}} \\
    Speech         & 17.9 & 32.5 & 22.9 & 10.9 \\
    Speech + Text  & 17.8 & 33.3 & 22.6 & 10.6 \\
    \midrule
    \multicolumn{5}{l}{\kern-0.5em\emph{Decoder pre-training}} \\
    Text-to-text & 9.6 & 26.4 & 13.6 & 2.3 \\
    Text-to-phoneme & 10.1 & 26.8 & 14.4 & 2.8 \\
    Phoneme-to-phoneme & 9.9 & 26.5 & 13.7 & 2.8 \\
    \midrule
    \multicolumn{5}{l}{\kern-0.5em\emph{Multi-task fine-tuning (with pre-trained mSLAM encoder)}} \\
    $\tau_\textrm{MT}=1.0$ & 19.1 & 34.2 & 24.2 & 12.0 \\
    $\tau_\textrm{MT}=5.0$ & 19.3 & 33.2 & 24.6 & 12.5 \\
    \midrule
    \multicolumn{5}{l}{\kern-0.5em\emph{TTS-based data aug. (with pre-trained mSLAM encoder)}} \\
    $\tau_\textrm{aug}=1.0$ & 19.7 & 33.9 & 25.5 & 12.6 \\
    $\tau_\textrm{aug}=5.0$ & 22.0 & 33.5 & 25.8 & 16.5 \\
    \midrule
    \emph{Reference} & 91.1 & 88.4 & 89.2 & 92.8 \\
    \bottomrule
\end{tabular}
\end{table}

\subsection{Baselines}
\label{exp:baselines}

As can be seen from the first group in Table~\ref{tbl:comparison},
replacing the LSTM-based linguistic decoder in the original Translatotron 2 with a Transformer-based decoder improved translation quality by +1.4 BLEU (or +16\% relatively) on average on all 21 language pairs, which is consistent with the general observations on the superior performance of Transformer in MT and ST tasks.

\subsection{Encoder pre-training}
\label{exp:encoder-pretraining}

We compared a w2v-BERT encoder pre-trained on unsupervised speech data and an mSLAM encoder pre-trained on semi-supervised speech and text data, as described in Sec.~\ref{sec:pre-trainng}. Both encoders used a 1024$\times$24 Conformer stack, with 0.6B parameters.
The results are shown in the second group in Table~\ref{tbl:comparison}. Both pre-trained encoders brought large boosts to the translation quality from the direct S2ST model (+7.8 BLEU, or +77\% relatively). Such improvement is shown across the board on low, mid and high-resource languages. These results suggest the effectiveness of pre-training on a massive amount of unlabeled speech data.

Consistent with ST experimental results from \citet{bapna2022mslam}, additional pre-training on a large amount of text data and a small amount of paired speech-transcript data did not further improve the performance of direct S2ST when it was directly used. However, as shown in Sec.~\ref{exp:multitask}, the speech-text joint representations from such pre-training is beneficial in multi-task fine-tuning.

\subsection{Decoder pre-training}
\label{exp:decoder-pretraining}

Compared to encoder pre-training in self-supervised tasks, decoder pre-training in supervised MT-derived  tasks was less effective. Experiments on three variants of MT tasks did not show any obvious gains, as shown in the third group in Table~\ref{tbl:comparison}. %

\subsection{Multimodal multi-task fine-tuning}
\label{exp:multitask}

As an alternative approach for utilizing MT data in S2ST, we trained a Translatotron 2 model and a Transformer MT model in a multi-task learning setup, with a pre-trained mSLAM encoder and a randomly initialized Transformer decoder shared between the two tasks (Figure~\ref{fig:multitask-finetuning}).
At each training step, one task was sampled randomly with equal probabilities and was trained with the corresponding data.
Because the MT dataset is extremely unbalanced among languages, we experimented with different data sampling temperature \citep{arivazhagan2019massively} on the MT task. 
As the results shown in the fourth group in Table~\ref{tbl:comparison}, such multimodal multi-task fine-tuning further improved the translation quality by +1.3 BLEU, suggesting that it is effective on untilizing weakly-supervised MT data for improving direct S2ST. There was no obvious performance difference by increasing the data sampling temperature $\tau_\textrm{MT}$ from 1.0 to 5.0.

\subsection{TTS-based data augmentation}

We used a multilingual multispeaker PnG NAT model \citep{jia2021png,shen2020non} to synthesize the source text from the MT dataset into speech, and used the same English TTS model as in \citet{jia2022cvss} to synthesize the targets. The multilingual PnG NAT model was trained on an proprietary dataset covering 20 source languages from CVSS-C except for Mongolian. For Mongolian text, we used a proprietary Mongolian G2P system to convert the text into phonemes and fed them into the multilingual PnG NAT together with the text written in the Cyrillic system, %
regardless that the model was not trained on any Mongolian data. 
The S2ST model is trained on a mixture of the CVSS-C corpus and the augmented dataset with equal data sampling probabilities.

As can been from the fifth group in Table~\ref{tbl:comparison}, training with the augmented dataset obtained higher performance improvement than the multimodal multi-task fine-tuning approach. In contrast to the latter, applying a high data sampling temperature $\tau_\textrm{aug}=5.0$ on the augmented dataset further boosted the performance significantly, especially on the low resource languages, which is consistent with increased data sampling ratio on them.
With $\tau_\textrm{aug}=5.0$ applied, the translation quality is further improved by +4.2 BLEU (or +24\% relatively) on average on top of the encoder pre-training.
The improvement on low-resource languages is drastic, e.g. 1.3$\to$18.5 on Indonesian, 7.0$\to$22.4 on Arabic.

\subsection{Scaling up}

We further examined scaling up the encoder and the decoder of Translatotron 2, specifically by increasing the encoder to a 1408$\times$32 Conformer with 1.8B parameters and the decoder to a 768$\times$12 Transformer with 113M parameters, following hyper-parameters from \citet{bapna2022mslam,vaswani2017attention}, respectively. 
The decoders were not pre-trained with MT tasks since no benefits were observed.%

The results are depicted in Table~\ref{tbl:scaling-up}. As can be seen, the impact of scaling up the encoder and the decoder is consistent among all approaches. Scaling up the encoder from 0.6B to 1.8B consistently further improved the performance by $\sim$3 BLEU in all groups; while scaling up the decoder from 25M to 113M brought only a minor improvement ($\le$ 0.3 BLEU). The best result was obtained by scaling up both the encoder and the decoder, alongside TTS-based augmentation, which outperformed the prior state-of-the-art trained on CVSS corpus only by +13.6 BLEU (or +113\% relatively). The improvement on the low resource languages are most significant (+398\% on average).

\begin{table}[t]
  \small
  \caption{Performance by scaling up the size of the S2ST model. All models used pre-trained mSLAM encoders.}
  \label{tbl:scaling-up}
  \centering
\begin{tabular}{rrrrrr}
    \toprule
    \multicolumn{2}{c}{\# Params} & \multicolumn{4}{c}{Avg BLEU} \\
    \cmidrule(r){1-2}\cmidrule(l){3-6}
    Encoder & Decoder & All & High & Mid& Low \\
    \midrule
    \multicolumn{3}{l}{\kern-0.5em\emph{Encoder pre-training}} \\
    0.6B &  25M & 17.8 & 33.3 & 22.6 & 10.6 \\
    0.6B & 113M & 18.1 & 33.7 & 22.8 & 10.9 \\
    1.8B & 25M & 20.2 & 33.9 & 25.7 & 13.3 \\
    \midrule
    \multicolumn{3}{l}{\kern-0.5em\emph{Multi-task fine-tuning ($\tau_\textrm{MT}=5.0$)}} \\
    0.6B & 25M & 19.3 & 33.2 & 24.6 & 12.5 \\
    1.8B & 113M  & 22.8 & 34.0 & 27.4 & 17.1 \\
    \midrule
    \multicolumn{6}{l}{\kern-0.5em\emph{TTS-based data aug. ($\tau_\textrm{aug}=5.0$)}} \\
    0.6B &  25M &  22.0 & 33.5 & 25.8 & 16.5  \\
    0.6B & 113M &  22.1 & 34.3 & 25.8 & 16.4 \\
    1.8B & 113M &  \B 25.6 & \B 35.1 & \B 29.4 & \B 20.9 \\
    \midrule
    \multicolumn{2}{l}{\emph{Prior state-of-the-art \citep{jia2022cvss}}} & 12.0 & 29.7 & 16.5 & 4.2 \\
    \midrule
    \multicolumn{2}{l}{\emph{Reference}} & 91.1 & 88.4 & 89.2 & 92.8 \\
    \bottomrule
    \vspace{-5ex}
\end{tabular}
\end{table}

\section{Discussion}

As the experimental results show, leveraging unsupervised and weakly-supervised data can drastically improve the performance of direct S2ST. Such improvements can be critical in practice because S2ST is a data scarce task. In our experiments, such improvements primarily came from self-supervised pre-training on a massive amount of unlabeled speech data and utilizing weakly-supervised MT data in the supervised S2ST training, both of which were effective across the board on high/mid/low-resource languages. We did not observe improvements from pre-training on an MT task and its variants.

Between the two approaches for utilizing MT data in supervised S2ST training that we explored, augmenting MT data into S2ST data by a TTS system obtained superior performance than S2ST and MT multimodal multi-task learning, when a pre-trained speech-text joint representation was utilized for the latter.

Despite the smaller improvement obtained from the multimodal multi-task learning approach, it has the advantage of lower computational overhead by avoiding synthesizing a large amount of text data into speech, and without concerns on potential bias towards synthetic speech input in the trained S2ST model.
These results confirmed the effectiveness of the speech-text joint representation learned from mSLAM, while also suggest headroom for further improvement.

While using a Transformer decoder obtained significant translation quality improvements,
because it includes cross-attention in every decoder layers, it may have a potential disadvantage on maintaining the temporal synchronization between the linguistic information and the acoustic information in the decoder output,  compared to the original Translatotron 2. Such a difference may negatively impact the performance on fine granularity paralinguistic and non-linguistic information preservation during speech translation, which we did not explore in this work. Possible mitigations include modifying the Transformer decoder to use cross-attention only in its bottom layer.

Because the baseline Translatotron 2 already obtained human-parity on the naturalness of the predicted translation speech on the CVSS-C corpus \citep{jia2022cvss}, its performance is unlikely bounded by the synthesizer component. Therefore, we did not explore utilizing more data for improving the synthesizer.

\section{Conclusions}

We explored multiple approaches for leveraging much more widely available unsupervised and weakly-supervised speech and text data to improve the performance of direct S2ST based on Translatotron 2.
With the most effective approaches applied, the translation quality of direct S2ST on the CVSS-C corpus was improved by +13.6 BLEU (or +113\% relatively) as the average on all the 21 language pairs, and +398\% relatively as the average on 12 low-resource language pairs, 
compared to the previous state-of-the-art trained without additional data. Our comparative studies suggest future research directions on S2ST and speech representation learning.

\section{Acknowledgments}

The authors thank Benjamin Lee for his help on improving the multi-task learning infrastructure in the Lingvo framework and Chung-Cheng Chiu for helpful feedback.

\newpage
\bibliographystyle{IEEEtran}
\bibliography{references}

\begin{thebibliography}{10}
\providecommand{\url}[1]{#1}
\csname url@samestyle\endcsname
\providecommand{\newblock}{\relax}
\providecommand{\bibinfo}[2]{#2}
\providecommand{\BIBentrySTDinterwordspacing}{\spaceskip=0pt\relax}
\providecommand{\BIBentryALTinterwordstretchfactor}{4}
\providecommand{\BIBentryALTinterwordspacing}{\spaceskip=\fontdimen2\font plus
\BIBentryALTinterwordstretchfactor\fontdimen3\font minus
  \fontdimen4\font\relax}
\providecommand{\BIBforeignlanguage}[2]{{%
\expandafter\ifx\csname l@#1\endcsname\relax
\typeout{** WARNING: IEEEtran.bst: No hyphenation pattern has been}%
\typeout{** loaded for the language `#1'. Using the pattern for}%
\typeout{** the default language instead.}%
\else
\language=\csname l@#1\endcsname
\fi
#2}}
\providecommand{\BIBdecl}{\relax}
\BIBdecl

\bibitem{jia2019direct}
Y.~Jia, R.~J. Weiss, F.~Biadsy, W.~Macherey, M.~Johnson, Z.~Chen, and Y.~Wu,
  ``Direct speech-to-speech translation with a sequence-to-sequence model,'' in
  \emph{Proc. Interspeech}, 2019.

\bibitem{kano2021transformer}
T.~Kano, S.~Sakti, and S.~Nakamura, ``Transformer-based direct speech-to-speech
  translation with transcoder,'' in \emph{Proc. SLT}, 2021.

\bibitem{jia2021translatotron}
Y.~Jia, M.~T. Ramanovich, T.~Remez, and R.~Pomerantz, ``Translatotron 2:
  High-quality direct speech-to-speech translation with voice preservation,''
  in \emph{Proc. ICML}, 2022.

\bibitem{tjandra2019speech}
A.~Tjandra, S.~Sakti \emph{et~al.}, ``Speech-to-speech translation between
  untranscribed unknown languages,'' in \emph{Proc. ASRU}, 2019.

\bibitem{zhang2020uwspeech}
C.~Zhang, X.~Tan, Y.~Ren, T.~Qin \emph{et~al.}, ``{UWSpeech}: Speech to speech
  translation for unwritten languages,'' in \emph{Proc. AAAI}, 2021.

\bibitem{lee2021direct}
A.~Lee, P.-J. Chen, C.~Wang, J.~Gu, X.~Ma \emph{et~al.}, ``Direct
  speech-to-speech translation with discrete units,'' in \emph{Proc. ACL},
  2022.

\bibitem{ma2021direct}
X.~Ma, H.~Gong, D.~Liu, A.~Lee, Y.~Tang, P.-J. Chen \emph{et~al.}, ``Direct
  simultaneous speech to speech translation,'' \emph{arXiv}, 2021.

\bibitem{lee2021textless}
A.~Lee, H.~Gong, P.-A. Duquenne, H.~Schwenk, P.-J. Chen \emph{et~al.},
  ``Textless speech-to-speech translation on real data,'' \emph{arXiv}, 2021.

\bibitem{jia2022cvss}
Y.~Jia, M.~T. Ramanovich \emph{et~al.}, ``{CVSS} corpus and massively
  multilingual speech-to-speech translation,'' in \emph{Proc. LREC}, 2022.

\bibitem{oord2018representation}
A.~v.~d. Oord, Y.~Li, and O.~Vinyals, ``Representation learning with
  contrastive predictive coding,'' \emph{arXiv}, 2018.

\bibitem{baevski2020wav2vec}
A.~Baevski, H.~Zhou, A.~Mohamed, and M.~Auli, ``wav2vec 2.0: A framework for
  self-supervised learning of speech representations,'' in \emph{Proc.
  NeurIPS}, 2020.

\bibitem{hsu2021hubert}
W.-N. Hsu, B.~Bolte, Y.-H.~H. Tsai, K.~Lakhotia \emph{et~al.}, ``{HuBERT}:
  Self-supervised speech representation learning by masked prediction of hidden
  units,'' \emph{TASLP}, 2021.

\bibitem{chung2021w2v}
Y.-A. Chung, Y.~Zhang, W.~Han, C.-C. Chiu, J.~Qin \emph{et~al.}, ``{w2v-BERT}:
  Combining contrastive learning and masked language modeling for
  self-supervised speech pre-training,'' in \emph{Proc. ASRU}, 2021.

\bibitem{wu2020self}
A.~Wu, C.~Wang, J.~Pino \emph{et~al.}, ``Self-supervised representations
  improve end-to-end speech translation,'' in \emph{Proc. Interspeech}, 2020.

\bibitem{li2020multilingual}
X.~Li, C.~Wang, Y.~Tang, C.~Tran, Y.~Tang, J.~Pino, A.~Baevski, A.~Conneau, and
  M.~Auli, ``Multilingual speech translation from efficient finetuning of
  pretrained models,'' in \emph{Proc. ACL}, 2021.

\bibitem{wang2021large}
C.~Wang, A.~Wu, J.~Pino, A.~Baevski, M.~Auli, and A.~Conneau, ``Large-scale
  self-and semi-supervised learning for speech translation,'' in \emph{Proc.
  Interspeech}, 2021.

\bibitem{anastasopoulos2021findings}
A.~Anastasopoulos, O.~Bojar, J.~Bremerman \emph{et~al.}, ``Findings of the
  {IWSLT} 2021 evaluation campaign,'' in \emph{Proc. IWSLT}, 2021.

\bibitem{babu2021xlsr}
A.~Babu, C.~Wang, A.~Tjandra \emph{et~al.}, ``{XLS-R}: Self-supervised
  cross-lingual speech representation learning at scale,'' \emph{arXiv}, 2021.

\bibitem{zheng2021fused}
R.~Zheng, J.~Chen, M.~Ma, and L.~Huang, ``Fused acoustic and text encoding for
  multimodal bilingual pretraining and speech translation,'' in \emph{Proc.
  ICML}, 2021.

\bibitem{bapna2021slam}
A.~Bapna, Y.-A. Chung, N.~Wu, A.~Gulati, Y.~Jia, J.~H. Clark, M.~Johnson
  \emph{et~al.}, ``{SLAM}: A unified encoder for speech and language modeling
  via speech-text joint pre-training,'' \emph{arXiv}, 2021.

\bibitem{bapna2022mslam}
A.~Bapna, C.~Cherry, Y.~Zhang \emph{et~al.}, ``{mSLAM}: Massively multilingual
  joint pre-training for speech and text,'' \emph{arXiv}, 2022.

\bibitem{wang2020covost2}
C.~Wang, A.~Wu, J.~Gu, and J.~Pino, ``{CoVoST 2} and massively multilingual
  speech translation,'' in \emph{Proc. Interspeech}, 2021.

\bibitem{berard2018end}
A.~B{\'e}rard, L.~Besacier \emph{et~al.}, ``End-to-end automatic speech
  translation of audiobooks,'' in \emph{Proc. ICASSP}, 2018.

\bibitem{bansal2018pre}
S.~Bansal, H.~Kamper, K.~Livescu, A.~Lopez, and S.~Goldwater, ``Pre-training on
  high-resource speech recognition improves low-resource speech-to-text
  translation,'' in \emph{Proc. NAACL}, 2019.

\bibitem{weiss2017sequence}
R.~J. Weiss, J.~Chorowski \emph{et~al.}, ``Sequence-to-sequence models can
  directly translate foreign speech,'' in \emph{Proc. Interspeech}, 2017.

\bibitem{anastasopoulos2018tied}
A.~Anastasopoulos and D.~Chiang, ``Tied multitask learning for neural speech
  translation,'' in \emph{Proc. NAACL}, 2018.

\bibitem{jia2019leveraging}
Y.~Jia, M.~Johnson, W.~Macherey, R.~J. Weiss, Y.~Cao, C.-C. Chiu \emph{et~al.},
  ``Leveraging weakly supervised data to improve end-to-end speech-to-text
  translation,'' in \emph{Proc. ICASSP}, 2019.

\bibitem{liu2019end}
Y.~Liu, H.~Xiong, Z.~He, J.~Zhang \emph{et~al.}, ``End-to-end speech
  translation with knowledge distillation,'' in \emph{Proc. Interspeech}, 2019.

\bibitem{duquenne2021multimodal}
P.-A. Duquenne \emph{et~al.}, ``Multimodal and multilingual embeddings for
  large-scale speech mining,'' in \emph{Proc. NeurIPS}, 2021.

\bibitem{gulati2020conformer}
A.~Gulati, J.~Qin, C.-C. Chiu, N.~Parmar, Y.~Zhang, J.~Yu, W.~Han, S.~Wang,
  Z.~Zhang \emph{et~al.}, ``Conformer: Convolution-augmented transformer for
  speech recognition,'' in \emph{Proc. Interspeech}, 2020.

\bibitem{vaswani2017attention}
A.~Vaswani, N.~Shazeer, N.~Parmar, J.~Uszkoreit, L.~Jones, A.~N. Gomez
  \emph{et~al.}, ``Attention is all you need,'' in \emph{Proc. NeurIPS}, 2017.

\bibitem{wang2021voxpopuli}
C.~Wang, M.~Rivi{\`e}re, A.~Lee, A.~Wu \emph{et~al.}, ``{VoxPopuli}: A
  large-scale multilingual speech corpus for representation learning,
  semi-supervised learning and interpretation,'' in \emph{Proc. ACL}, 2021.

\bibitem{ardila2020common}
R.~Ardila, M.~Branson, K.~Davis \emph{et~al.}, ``{Common Voice}: A
  massively-multilingual speech corpus,'' in \emph{Proc. LREC}, 2020.

\bibitem{pratap2020mls}
V.~Pratap, Q.~Xu, A.~Sriram \emph{et~al.}, ``{MLS}: A large-scale multilingual
  dataset for speech research,'' in \emph{Proc. Interspeech}, 2020.

\bibitem{gales2014speech}
M.~J. Gales, K.~M. Knill, A.~Ragni, and S.~P. Rath, ``Speech recognition and
  keyword spotting for low-resource languages: Babel project research at
  cued,'' in \emph{Proc. SLTU}, 2014.

\bibitem{joshi2020spanbert}
M.~Joshi, D.~Chen, Y.~Liu \emph{et~al.}, ``{SpanBERT}: Improving pre-training
  by representing and predicting spans,'' \emph{TACL}, 2020.

\bibitem{graves2006connectionist}
A.~Graves, S.~Fern{\'a}ndez, F.~Gomez, and J.~Schmidhuber, ``Connectionist
  temporal classification: labelling unsegmented sequence data with recurrent
  neural networks,'' in \emph{Proc. ICML}, 2006.

\bibitem{conneau2019cross}
A.~Conneau and G.~Lample, ``Cross-lingual language model pretraining,'' in
  \emph{Proc. NeurIPS}, 2019.

\bibitem{xue2020mt5}
L.~Xue, N.~Constant, A.~Roberts, M.~Kale, R.~Al-Rfou, A.~Siddhant, A.~Barua,
  and C.~Raffel, ``{mT5}: A massively multilingual pre-trained text-to-text
  transformer,'' in \emph{Proc. NAACL}, 2021.

\bibitem{barrault2020findings}
L.~Barrault, M.~Biesialska, O.~Bojar, M.~R. Costa-juss{\`a} \emph{et~al.},
  ``Findings of the 2020 conference on machine translation ({WMT20}),'' in
  \emph{Proc. Conference on Machine Translation}, 2020.

\bibitem{barrault2019findings}
L.~Barrault, O.~Bojar, M.~R. Costa-Jussa, C.~Federmann \emph{et~al.},
  ``Findings of the 2019 conference on machine translation ({WMT19}),'' in
  \emph{Proc. Conference on Machine Translation}, 2019.

\bibitem{bojar2018findings}
O.~Bojar, C.~Federmann, M.~Fishel, Y.~Graham \emph{et~al.}, ``Findings of the
  2018 conference on machine translation ({WMT}18),'' in \emph{Proc. Conference
  on Machine Translation}, 2018.

\bibitem{bojar2017findings}
O.~Bojar, R.~Chatterjee, C.~Federmann, Y.~Graham \emph{et~al.}, ``Findings of
  the 2017 conference on machine translation ({WMT}17),'' in \emph{Proc.
  Conference on Machine Translation}, 2017.

\bibitem{bojar2015findings}
O.~Bojar, R.~Chatterjee, C.~Federmann, B.~Haddow \emph{et~al.}, ``Findings of
  the 2015 workshop on statistical machine translation,'' in \emph{Proc.
  Workshop on Statistical Machine Translation}, 2015.

\bibitem{bojar2013findings}
O.~Bojar, C.~Buck, C.~Callison-Burch, C.~Federmann \emph{et~al.}, ``Findings of
  the 2013 {W}orkshop on {S}tatistical {M}achine {T}ranslation,'' in
  \emph{Proc. Workshop on Statistical Machine Translation}, 2013.

\bibitem{qi2018and}
Y.~Qi, D.~S. Sachan, M.~Felix, S.~J. Padmanabhan, and G.~Neubig, ``When and why
  are pre-trained word embeddings useful for neural machine translation?'' in
  \emph{Proc. NAACL}, 2018.

\bibitem{kirkpatrick2017overcoming}
J.~Kirkpatrick, R.~Pascanu, N.~Rabinowitz, J.~Veness \emph{et~al.},
  ``Overcoming catastrophic forgetting in neural networks,'' \emph{PNAS}, 2017.

\bibitem{shen2019lingvo}
J.~Shen, P.~Nguyen, Y.~Wu \emph{et~al.}, ``Lingvo: A modular and scalable
  framework for sequence-to-sequence modeling,'' \emph{arXiv}, 2019.

\bibitem{park2020improved}
D.~S. Park, Y.~Zhang, Y.~Jia \emph{et~al.}, ``Improved noisy student training
  for automatic speech recognition,'' in \emph{Proc. Interspeech}, 2020.

\bibitem{arivazhagan2019massively}
N.~Arivazhagan \emph{et~al.}, ``Massively multilingual neural machine
  translation in the wild: Findings and challenges,'' \emph{arXiv}, 2019.

\bibitem{jia2021png}
Y.~Jia, H.~Zen \emph{et~al.}, ``{PnG BERT}: Augmented {BERT} on phonemes and
  graphemes for neural {TTS},'' in \emph{Proc. Interspeech}, 2021.

\bibitem{shen2020non}
J.~Shen, Y.~Jia, M.~Chrzanowski, Y.~Zhang, I.~Elias, H.~Zen \emph{et~al.},
  ``Non-{A}ttentive {T}acotron: Robust and controllable neural {TTS} synthesis
  including unsupervised duration modeling,'' \emph{arXiv}, 2020.

\end{thebibliography}

\newpage
\appendix
\onecolumn

\begin{landscape}

\section{Detailed evaluation results of the S2ST models}
\label{a:eval-details}

\begin{table*}[h]
\centering
\begin{small}
\footnotesize
\caption{Performance of the multilingual X$\to$En S2ST models on CVSS-C test sets. Evaluated by BLEU on ASR transcription.}
\setlength{\tabcolsep}{0.45em}
\begin{tabular}{lrrrrrrrrrrrrrrrrrrrrrrrr}
    \toprule
     & \multicolumn{2}{c}{\#Params} & & \multicolumn{4}{c}{High-resource} & \multicolumn{5}{c}{Mid-resource} & \multicolumn{12}{c}{Low-resource} \\
     \cmidrule{2-3}\cmidrule(r){5-8}\cmidrule(lr){9-13}\cmidrule(l){14-25}
    X$\to$En & Encoder & Decoder & Avg & fr & de & ca & es & fa & it & ru & zh & pt & nl & tr & et & mn & ar & lv & sl & sv & cy & ta & ja & id \\
    Hours (source) & & & & 264 & 184 & 136 & 113 & 49 & 44 & 18 & 10 & 10 & 7.3 & 4.1 & 3.4 &  3.0 &  2.1 &  2.1 &  2.0 &  1.7 &  1.7 &  1.6 & 1.3 & 1.2  \\
    
    \midrule
    \multicolumn{4}{l}{\kern-0.5em\emph{From scratch}} \\
    Baselines \citep{jia2022cvss} & 26M & 10M & 8.7 & 28.3 & 19.7 & 23.5 & 30.1 & 2.4 & 24.1 & 19.6 & 4.5 & 12.5 & 6.5 & 3.8 & 0.6 & 0.2 & 1.7 & 1.5 & 0.4 & 1.3 & 0.9 & 0.1 & 0.5 & 0.4 \\
    This work & 26M & 25M & 10.1 & 29.5 & 22.3 & 25.0 & 30.8 &  3.4 & 26.0 & 21.7 &  5.5 & 14.3 & 10.5 &  6.6 &  1.1 &  0.2 &  3.8 &  3.0 &  2.3 &  2.8 &  1.6 &  0.1 &  0.5 &  0.8 \\
    \midrule
    \multicolumn{4}{l}{\kern-0.5em\emph{Encoder pre-training}} \\
    Speech & 0.6B & 25M & 17.9 & 33.6 & 30.6 & 30.1 & 35.9 &  6.0 & 32.5 & 38.9 &  5.2 & 31.9 & 29.3 &  9.2 & 16.0 &  0.2 & 10.4 & 15.6 & 17.8 & 25.9 &  4.2 &  0.3 &  0.9 &  1.5 \\
    Speech + Text & 0.6B & 25M & 17.8 & 34.5 & 30.7 & 31.1 & 36.9 &  5.9 & 33.8 & 38.1 &  4.0 & 31.1 & 28.6 &  8.1 & 15.0 &  0.2 &  7.0 & 16.0 & 18.6 & 28.1 &  3.5 &  0.2 &  0.9 &  1.3 \\
    Speech + Text & 0.6B & 113M & 18.1 & 34.9 & 31.1 & 31.7 & 37.0 &  4.6 & 34.0 & 38.3 &  3.8 & 33.3 & 30.6 &  7.1 & 15.0 &  0.2 &  5.8 & 17.4 & 20.8 & 29.0 &  2.7 &  0.3 &  0.8 &  1.1 \\
    Speech + Text & 1.8B & 25M & 20.2 & 35.4 & 31.6 & 31.6 & 36.9 & 10.4 & 34.8 & 40.1 &  3.3 & 40.1 & 32.2 & 17.3 & 16.9 &  0.1 &  9.9 & 21.6 & 24.7 & 33.4 &  1.6 &  0.1 &  0.5 &  1.0 \\
    \midrule
    \multicolumn{4}{l}{\kern-0.5em\emph{Decoder pre-training}} \\
    Text-to-text & 26M & 25M & 9.6 & 28.9 & 21.6 & 24.5 & 30.7 &  2.8 & 25.0 & 21.4 &  5.6 & 13.1 &  8.6 &  6.2 &  1.0 &  0.2 &  2.8 &  1.9 &  1.7 &  1.5 &  1.6 &  0.2 &  0.6 &  0.8 \\
    Text-to-phoneme & 26M & 25M & 10.1 & 29.5 & 22.0 & 25.3 & 30.5 &  3.8 & 25.3 & 22.4 &  4.9 & 15.6 & 10.5 &  7.3 &  0.8 &  0.2 &  4.1 &  2.6 &  3.2 &  2.2 &  1.0 &  0.1 &  0.8 &  1.0 \\
    Phoneme-to-phoneme & 26M & 25M & 9.9 & 29.2 & 22.0 & 24.5 & 30.4 &  3.7 & 25.1 & 22.4 &  5.6 & 11.7 &  9.8 &  7.3 &  1.0 &  0.2 &  4.0 &  2.1 &  2.3 &  3.0 &  1.3 &  0.2 &  0.7 &  1.4 \\
    \midrule
    \multicolumn{4}{l}{\kern-0.5em\emph{Multi-task fine-tuning}} \\
    $\tau_\textrm{MT}=1.0$ & 0.6B & 25M & 19.1 & 35.3 & 32.0 & 31.7 & 37.7 &  6.9 & 34.5 & 41.5 &  5.1 & 33.1 & 31.0 & 10.9 & 16.6 &  0.2 &  7.9 & 17.3 & 23.7 & 31.9 &  2.7 &  0.1 &  0.7 &  1.2 \\
    $\tau_\textrm{MT}=5.0$ & 0.6B & 25M & 19.3 & 33.9 & 31.5 & 30.6 & 36.8 &  7.2 & 33.7 & 41.6 &  6.4 & 34.1 & 31.1 & 16.1 & 17.1 &  0.3 & 10.0 & 14.4 & 22.9 & 28.4 &  5.4 &  0.2 &  1.3 &  2.5 \\
    $\tau_\textrm{MT}=5.0$ & 1.8B & 113M & 22.8 & 34.8 & 32.7 & 31.5 & 36.8 &  9.6 & 34.4 & 44.9 &  7.2 & 40.8 & 34.1 & 24.9 & 21.5 &  0.5 & 20.1 & 21.3 & 26.3 & 34.1 &  4.6 &  0.4 &  1.5 & 16.0 \\
    \midrule
    \multicolumn{4}{l}{\kern-0.5em\emph{TTS-based data aug.}} \\
    $\tau_\textrm{aug}=1.0$ & 0.6B & 25M & 19.7 & 35.0 & 32.2 & 31.1 & 37.2 &  5.9 & 33.8 & 43.2 & 11.4 & 33.0 & 29.8 & 11.7 & 16.6 &  0.4 & 11.2 & 15.5 & 22.7 & 30.1 &  4.0 &  0.1 &  5.7 &  2.9 \\
    $\tau_\textrm{aug}=1.0$ & 1.8B & 113M & 23.5 & 36.0 & 33.7 & 31.8 & 38.2 &  8.2 & 35.3 & 46.1 & 15.0 & 39.4 & 33.5 & 23.4 & 20.8 &  0.2 & 19.2 & 21.9 & 26.3 & 33.5 &  3.4 &  0.3 &  9.5 & 18.2 \\
    $\tau_\textrm{aug}=5.0$ & 0.6B & 25M & 22.0 & 34.5 & 32.0 & 30.7 & 37.1 &  8.2 & 33.8 & 42.6 & 10.6 & 34.0 & 31.8 & 23.9 & 17.2 &  1.1 & 22.4 & 15.6 & 23.3 & 31.1 &  7.6 &  0.6 &  5.5 & 18.5 \\
    $\tau_\textrm{aug}=5.0$ & 0.6B & 113M & 22.1 & 35.7 & 32.0 & 31.6 & 37.8 &  5.9 & 35.0 & 42.4 &  9.0 & 36.9 & 32.0 & 21.0 & 16.9 &  0.6 & 19.5 & 18.5 & 25.4 & 32.7 &  4.3 &  0.4 &  4.0 & 22.0 \\
    $\tau_\textrm{aug}=5.0$ & 1.8B & 113M & 25.6 & 36.5 & 33.6 & 31.9 & 38.5 & 11.6 & 35.7 & 45.6 & 13.1 & 41.1 & 34.1 & 28.7 & 21.0 &  2.5 & 30.2 & 22.7 & 25.8 & 36.6 &  5.4 &  2.2 &  8.5 & 32.8 \\

    \midrule
    \emph{Prior state-of-the-art \citep{jia2022cvss}} & 26M & 10M & 12.0 & 32.4 & 24.8 & 28.2 & 33.4 & 6.3 & 28.6 & 23.2 & 6.3 & 18.3 & 15.8 & 10.6 & 2.5 & 0.4 & 5.4 & 2.3 & 3.1 & 3.2 & 4.5 & 0.1 & 1.0 & 1.0 \\
    
    \midrule
    \emph{Reference} & & & 91.1 & 84.6 & 88.4 & 92.0 & 88.6 & 91.7 & 89.5 & 94.0 & 77.8 & 93.1 & 90.6 & 92.7 & 89.3 & 92.4 & 94.2 & 94.8 & 94.9 & 94.1 & 92.0 & 90.6 & 95.3 & 92.6 \\
    \bottomrule
\end{tabular}
\label{tbl:eval-details}
\end{small}
\end{table*}

\newpage

\section{Amount of the MT dataset and the TTS-augmented S2ST dataset}
\label{a:data-details}

\begin{table*}[h]
\centering
\begin{small}
\footnotesize
\caption{Amount of source data in the MT dataset and the TTS-augmented S2ST dataset.}
\setlength{\tabcolsep}{0.45em}
\begin{tabular}{lrrrrrrrrrrrrrrrrrrrrrc}
    \toprule
     & \multicolumn{4}{c}{High-resource} & \multicolumn{5}{c}{Mid-resource} & \multicolumn{12}{c}{Low-resource} & \multirow{2.5}{*}{Total} \\
     \cmidrule(r){2-5}\cmidrule(lr){6-10}\cmidrule(l){11-22}
     & fr & de & ca & es & fa & it & ru & zh & pt & nl & tr & et & mn & ar & lv & sl & sv & cy & ta & ja & id \\
    \midrule
    MT ($\times 10^3$ sentences) & 38,288 & 38,363 & 96 & 13,134 & 205 & 236 & 33,512 & 22,200 & 61 & 191 & 210 & 2,152 & 10 & 216 & 623 & 22 & 59 & 1 & 738 & 17,879 & 89 & 168,285 \\
    S2ST (hours) & 103,157 & 82,037 & 97 & 37,240 & 397 & 397 & 81,637 & 46,511 & 92 & 286 & 539 & 3,796 & 13 & 443 & 1,553 & 34 & 80 & 1 & 1,413 & 39,102 & 156 & 398,983 \\
    \bottomrule
\end{tabular}
\label{tbl:data-details}
\end{small}
\end{table*}

\end{landscape}

\end{document}